\documentclass[10pt,twocolumn,letterpaper]{article}

\usepackage[square,sort,comma,numbers]{natbib}
\usepackage{graphicx}
\usepackage{dcolumn}
\usepackage{bm}
\date{}
\begin{document}


\title{Sales forecasting using WaveNet within the framework of the Kaggle competition}

\author{Glib Kechyn, Lucius Yu, Yangguang Zang, Svyatoslav Kechyn}

\maketitle

\section{\label{sec:level1}Abstract}

We took part in the Corporacion Favorita Grocery Sales Forecasting competition\cite{kaggle_competition} hosted on Kaggle and achieved the 2nd place. In this abstract paper, we present an overall analysis and solution to the underlying machine-learning problem based on time series data, where major challenges are identified and corresponding preliminary methods are proposed. Our approach is based on the adaptation of dilated convolutional neural network for time series forecasting. By applying this technique iteratively to batches of n examples, a big amount of time series data can be eventually processed with a decent speed and accuracy. We hope this paper could serve, to some extent, as a review and guideline of the time series forecasting benchmark, inspiring further attempts and researches.

\section{Introduction}

Time series competitions offered by Kaggle and other organizations have become popular of machine learning. By initiating common rules of participation as well as training and testing datasets that are shared by all contestants, these competitions can help to advance the state-of-the-art of machine learning practice in a variety of application domains.
Time Series is considered to be one of the less known skills in the analytics space. Seasonality, trends and cycles exist in data and it is difficult to identify and predict accurately due to the non-linear trends and noise presented in the series. The significant increase of neural networks popularity has given a radically different understanding how forecasting could be done. Advances in hardware have made it possible to tackle problems with deep neural nets in a reasonable amount of time. Now that it is a feasible solution, deep learning has set a lot of new records for accurate classification on benchmark datasets in recent years \cite{ching2018opportunities, shvets2018automatic}. 
In this paper, we discuss our approach to solving this Kaggle challenge: Corporacion Favorita Grocery Sales Forecasting. We describe and analyze applying Convolutional Neural Networks in the context of time series data.

\section{Dataset Description}

According to Kaggle competitions format, the data is split into two types - train data and test data. Train data represents data for model training while test data is split into parts and used for model's accuracy evaluation on public and private leaderboards.
Corporacion Favorita consists of 125,497,040 observations in train and 3,370,464 in test. Datasets were composed of sales by date, store number, item number, and promotion information. Besides, transactions information, oil prices, store information and holidays days were provided as well.

\section{Evaluation Metric}

The competition uses NWRMSLE (normalized weighted root mean squared logarithmic error) as the evaluation metrics.
\begin{equation}
    \sqrt{\frac{\sum_{i=1}^{n}wi(ln(\hat{y}_{i}+1) - ln(y_{i}+1))^{2}}{\sum_{i=1}^{n}wi}}
\end{equation}

Where for row i,  ̂\begin{equation}yiy^i \end{equation}
is the predicted sales of an item and yiyi is the actual sales; n is the total number of rows in the test set.
The weights, wiwi, can be found in the dataset. Perishable items are given a weight of 1.25 where all other items are given a weight of 1.00.
\begin{figure}[h]
    \includegraphics[scale=0.35]{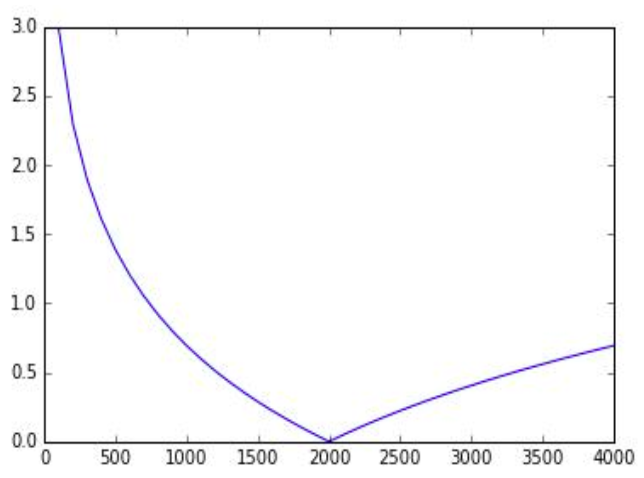}
    \caption{Metric visualization.}
    \label{fig:metric}
\end{figure}
This metric is suitable when predicting values across a large range of orders of magnitudes. It avoids penalizing large differences in prediction when both the predicted and the true number are large: predicting 5 when the true value is 50 is penalized more than predicting 500 when the true value is 545.
It can be seen on \label{fig:metric} Figure 1, whereby the left side is true positive values and number of iterations is in the bottom. 

\section{Problem Definition}

Brick-and-mortar grocery stores are always in a delicate dance with purchasing and sales forecasting. Predict a little over, and grocers are stuck with overstocked, perishable goods. Guess a little under, and popular items quickly sell out, leaving money on the table and customers fuming.
The problem becomes more complex as retailers add new locations with unique needs, new products, ever transitioning seasonal tastes, and unpredictable product marketing.

Kaggle community has been challenged to build a model that forecasts product sales more accurately.  In other words, our goal is to build a model with the highest accuracy for sales forecasting, with a possibility of using the model in production.

\section{Challenges}
The defined forecasting problem has at least the following challenges:
\begin{enumerate}
   \item Noisy data:
    while the organizers tried as best as possible to prepare data for participants and collected a large amount of data, the problems with noisy labels existed. Some of the data (oil prices, holidays, transactions) did not correlate with the target and was not used in the future.
    \item Unseen data:
    there was such a situation where unseen data occurred in the test dataset. This means that the model behaves unpredictably with respect to unseen store/item data.The reason for this is that the training set does not include records for zero sales. The test set, though, includes all store/item combinations, whether or not that item was seen previously in a store. Finally, those combinations were replaced by zeroes with an assumption that unseen combinations were just zero sales data.
    \item Accuracy:
    since this experiment was conducted within the framework of the competition, all the possibilities were used to increase the accuracy of predictions.
    \end{enumerate}
\section{Alternative Approaches}
The architectures which are presented below, in terms of neural networks, have not performed better than our final WaveNet\cite{wavenet} model, in context of the current challenge, but we believe that they can provide interesting insights of problem-solving in a different way or can overperform WaveNet in the context of the other competitions.

\subsection* {Recurrent Neural Networks}

Recurrent neural networks\cite{rnn_common} can use the output of the current node as the input for the next node, which can be viewed as follows:

\includegraphics[scale=0.4]{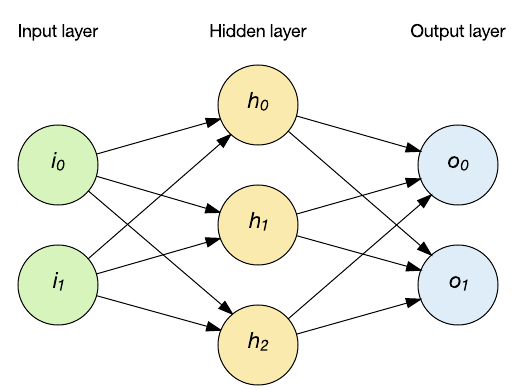}
Between the others Deep Learning algorithms, RNN has been commonly used in forecasting time series applications\cite{drnn} and is positioned as a state-of-the-art method for forecasting such data arrays.
The efficiency of these networks can be explained by the recurrent connections that allow the network to access the entire history of previous time series values.
 
A recurrent neural network\cite{rnn_c_r} can be thought of as multiple copies of the same network, each passing a message to a successor. Due to their nature, recurrent neural networks are intimately related to sequences and lists. In the last few years, there has been incredible success applying RNNs to a time series forecasting problems.

Significant to these successes is the use of "LSTMs"\cite{lstm}, a very special kind of recurrent neural network which works, for many tasks, much better than the standard RNN version. Almost all exciting results based on recurrent neural networks are achieved with them.

In such cases, where we need to look at recent information to perform the present task and the gap between the relevant information and the place that it's needed is small, RNNs can learn to use the past information. But there are also cases where we need more context. Unfortunately, as that gap grows, standard implementations of RNNs become unable to learn to connect the information. In these cases, LSTM networks fit as well as possible.

LSTM networks designed to avoid long term relationships\cite{long_term, lstm_forecasting}.

\begin{figure*}
  \includegraphics[width=\textwidth]{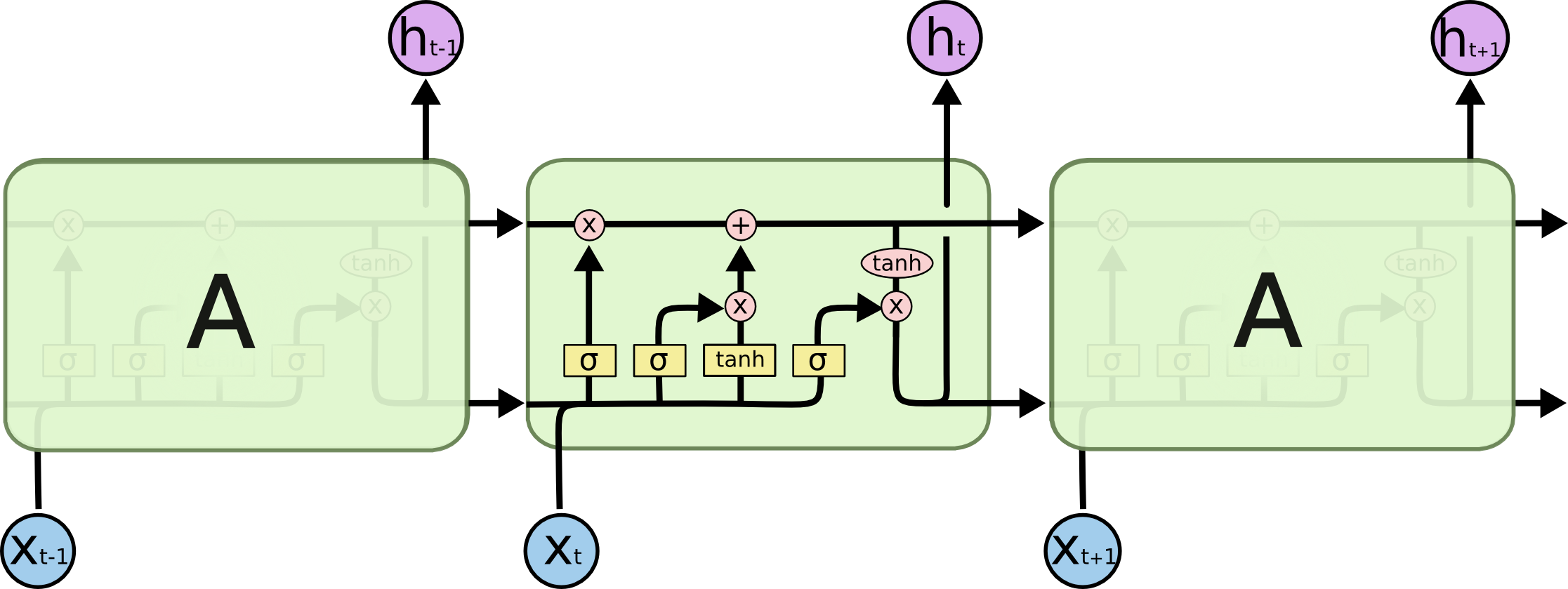}
  \caption{LSTM}
  \includegraphics[scale=0.45]{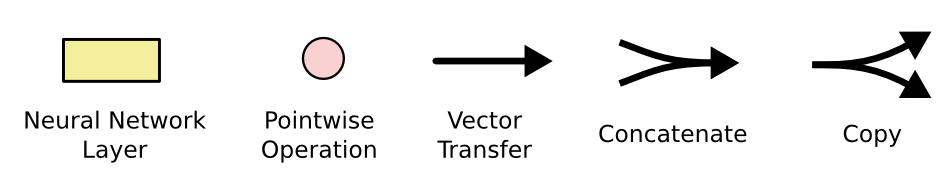}
\end{figure*}

\begin{figure*}
  \includegraphics[width=\textwidth]{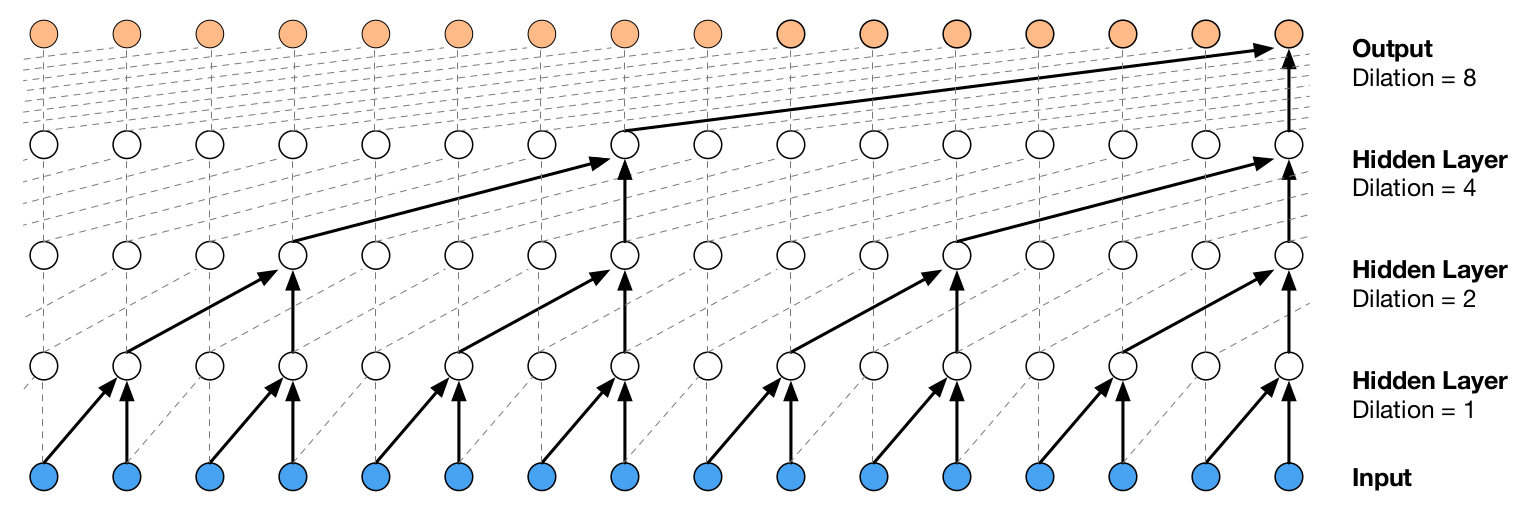}
  \label{dilation}
  \caption{Dilation convolution}
\end{figure*}

LSTM architecture assume having units with including a 'memory cell' (Figure 2) that can maintain information in memory for long periods of time. A set of gateways is used to control information in-out memory flow. On this basis, this architecture considered as state-of-the-art forecasting method.

Architectures as GRU\cite{gru} could also be used as a workaround to current problem as they are similar to LSTMs, but use a simplified structure.
\section{The Winning Approach}
\begin{figure}
    \includegraphics[scale=0.4]{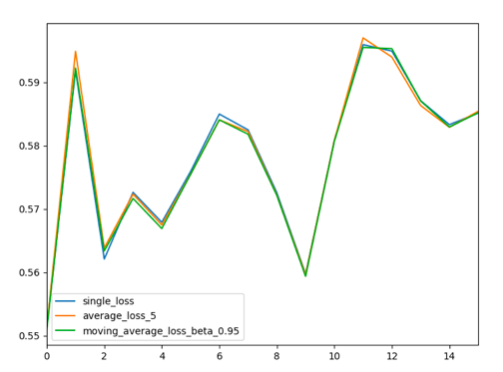}
    \caption{Day by day loss.}
\end{figure}
Our general pipeline follows the prototype WaveNet CNN model\cite{sjv}, with extensions and modifications\cite{enc_dec}.

The recent achievements of deep learning techniques motivated researchers to explore methods and techniques such as WaveNet in time series forecasting areas.\cite{wavenet_financtial}
WaveNet is a generative model. This means that the model can generate the sequences of real-valued data to some conditional inputs. The main idea behind the architecture is dilated causal convolutions (Figure 3).
With a lack of recurrent connections and skipping steps, causal convolutions may be considered faster to train than RNN. Big amount of layers or large filters used for increasing the receptive field is one of the problems of causal convolutions. Dilated convolutions solve these problems by using upsampled filters instead of feature maps. In other words, dilated convolutions allow you to preserve the size of the feature maps from one layer to another by only increasing the field of view of the kernel otherwise capture the global view of the input with fewer parameters.

To have a possibility to generate predictions for 16 days (forecasting target period) the model was modified. 
Since training was using next step predictions, errors accumulated. To handle this the sequence to sequence\cite{seq_to_seq} approach was used where the encoder and decoder\cite{enc_dec} did not share parameters. This allows the decoder to handle the accumulating noise when generating long sequences. Adam optimizer was used to update network weights.
\begin{figure}{b}
    \includegraphics[scale=0.4]{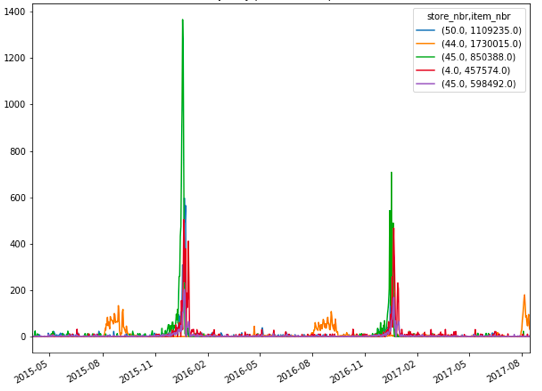}
    \caption{Trends and cycles.}
\end{figure}
\begin{figure}
    \includegraphics[scale=0.4]{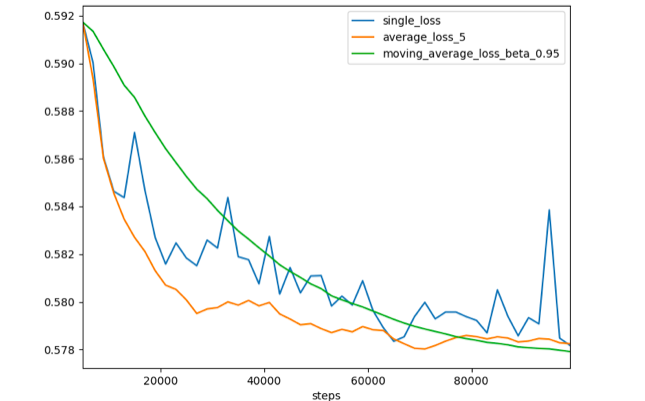}
    \caption{Moving average vs single model predictions.}
\end{figure}
Data was fed by mini-batches with randomly sampled 128 sequences. Then start of decode (target) date was chosen randomly. Since the whole dataset is around 170,000 (seq) x 365 days, the model mostly sees different data for each training iteration. Considering this, the model is good at handling overfitting.
During the training, learning rate was being set at 0.0005 with decay factor.

Validation was organized in step by step way. We kept the last 16 days of train data as a holdout set and used it for validation. 
Shifted sales and onpromotion data were used to capture better quarterly and yearly pattern.

The prediction from the model will be slightly overfitted during the last several hundreds or thousands of mini-batches. So the result has a little fluctuation.
To handle overfitting and to smooth out short-term fluctuations and highlight longer-term trends and cycles - moving average was used. Our model started to predict after 5000 mini-batch iterations and generated predictions every 2000 iterations. After certain iteration, as you can see in Figure 5, an average of 5 models performed better than a single model. To increase accuracy more, finally, an exponential moving average was used with a smooth factor which was calculated on local cross-validation.

With regard to features, the model has a good capability on catching the patterns of time series data so not a lot of features were used in it.
Some of them are unit sales and onpromotion information, unit sales shifted to the past and onpromotion information shifted to the future and past.

\section{Conclusion}
The obvious problem facing every business is that markets are unpredictable. Any sales forecasting, however rigorous its analysis of conditions, can be flat-out wrong. If market conditions remain relatively unchanged, a reliable method of forecasting is using historical data. \cite{forecasting_risk}

Our experience shows that convolutional neural networks are extremely good at handling historical data and catching seasonality, trends, cycles, and irregular components as shown in Figure 4.

We described an approach of using CNN WaveNet, a sequence to sequence architecture, in terms of sales forecasting,  which showed itself to be a highly-effective method (Figure 6) of solving time series predicting problems. Besides, we briefly described some other state-of-the-art techniques which could also be used as solutions.

As future work, CNN architectures with a greater amount of layers should be investigated for more difficult tasks. A big amount of data is required in order to train a deeper architecture. Therefore, modifying the pipeline for different types of data and domains can be another interesting future direction. Furthermore, exploring different ensemble techniques could give a decent accuracy gain as well.
To conclude, this paper could be considered as an abstract review of the application of deep learning technique to time series forecasting tasks.

\medskip
\bibliography{references}
\bibliographystyle{unsrt}

\end{document}